# DEEP LEARNING-BASED AVERAGE SHEAR WAVE VELOCITY PREDICTION USING ACCELEROMETER RECORDS

B. Yilmaz[1], M. Turkmen[1], S. Meral[2], E. Akagündüz[1], S. Tileylioglu[3]

[1] Dept. of Modeling and Simulation, Graduate School of Informatics, METU, Ankara, Turkey, e265058@metu.edu.tr

[2] Turkish Aerospace Industries, Inc., Ankara, Turkey

[3] Dept. of Civil Engineering, Kadir Has University, Istanbul, Turkey

**Abstract**: Assessing seismic hazards and thereby designing earthquake-resilient structures or evaluating structural damage that has been incurred after an earthquake are important objectives in earthquake engineering. Both tasks require critical evaluation of strong ground motion records, and the knowledge of site conditions at the earthquake stations plays a major role in achieving the aforementioned objectives. Site conditions are generally represented by the time-averaged shear wave velocity in the upper 30 meters of the geological materials ($Vs_{30}$). Several strong motion stations lack $Vs_{30}$ measurements resulting in potentially inaccurate assessment of seismic hazards and evaluation of ground motion records. In this study, we present a deep learning-based approach for predicting $Vs_{30}$ at strong motion station locations using three-channel earthquake records. For this purpose, Convolutional Neural Networks (CNNs) with dilated and causal convolutional layers are used to extract deep features from accelerometer records collected from over 700 stations located in Turkey. In order to overcome the limited availability of labeled data, we propose a two-phase training approach. In the first phase, a CNN is trained to estimate the epicenters, for which ground truth is available for all records. After the CNN is trained, the pre-trained encoder is fine-tuned based on the $Vs_{30}$ ground truth. The performance of the proposed method is compared with machine learning models that utilize hand-crafted features. The results demonstrate that the deep convolutional encoder based $Vs_{30}$ prediction model outperforms the machine learning models that rely on hand-crafted features. This suggests that our computational model can extract meaningful and informative features from the accelerometer records, enabling more accurate $Vs_{30}$ predictions. The findings of this study highlight the potential of deep learning-based approaches in seismology and earthquake engineering.

## 1. Introduction

Seismic events are monitored and recorded through seismic networks to better understand their impact on the built environment and hence to improve existing codes and enhance the safe design of structures. The primary product of seismic monitoring is the seismic catalog, which consists of a list of all earthquakes, explosions, and other seismic events, both natural and human-induced (National Research Council, 2006). Seismograph networks are distributed across different parts of the world to detect weak ground motions and are preferably installed on firm bedrock conditions to minimize variable site conditions between stations (Molnar et al., 2017). These networks are used to calculate the earthquake epicenter, magnitude, and depth. Strong ground motion stations, which are the focus of this study, are installed close to the earthquake source and on different types of soils (Molnar et al., 2017). Installing strong ground motion stations in close proximity to the source and on different soil types allows for the investigation of both how the soil affects seismic waves and its impact on structures that are situated on different soils. The growing quantity and diversity of data has led to significant advancements in identifying local ground effects, ground motion prediction equations, and seismic hazard analysis. An important aspect of soundly interpreting and analyzing earthquake motions recorded at various sites involves knowledge of field conditions at strong-motion stations.

Seismic hazard analysis is typically conducted using ground motion prediction equations, which have been developed by many researchers through regression analyses utilizing data from the strong ground motion databases. Ground motion prediction equations take into account various factors, including earthquake magnitude, distance, and local site effects. The time-averaged shear wave velocity to a depth of 30 meters depth from the surface ($Vs_{30}$) holds critical importance in seismic risk analysis and the assessment of local ground conditions. Seismic waves propagate at different velocities in various soil types beneath the Earth's surface. Higher $Vs_{30}$ values generally indicate firmer and stiffer soils, while lower $Vs_{30}$ values suggest softer and more elastic soils. In the data sets created for developing ground motion prediction equations, recordings from strong motion stations with known $Vs_{30}$ values are often used. In cases where geophysical tests aren't available for for certain stations, the values are estimated through proxies such as site topography (Wald and Allen, 2007) and site frequency (Hassani and Atkinson, 2016) which may be calculated using the Horizontal-to-Vertical Spectral Ratio (HVSR) method applied onto microtremor measurements or strong motion records. There are methods that attempt to directly estimate $Vs_{30}$ from earthquake recordings. Recently, Ni et al. (2014) and Kim et al. (2016) presented a method to estimate $Vs_{30}$ by modeling the horizontal-to-vertical ratio of local P-wave seismograms. The success of the method however, is dependent on how well the crustal velocity of the region is known.

The studies summarized so far predominantly utilize correlations and the results are rather than the utilization of contemporary Artificial Intelligence AI techniques that are driven by data. In the following sections, we first summarize previous studies that have used AI techniques to predict $Vs_{30}$ at strong motion station sites. We then present the data set used in this study, followed by our methodology and results. The paper concludes with an in-depth examination of our findings, accompanied by future potential research directions.

## 2. Previous Studies Using AI to Estimate $Vs_{30}$

The amount of earthquake data gathered from seismic events has grown considerably in recent years (Alimoradi A., Beck J.L., 2014; Mosher, S. G., & Audet, P., 2020). As a result, there are now opportunities to utilize AI techniques to analyze earthquake data. There has been, however, limited research in identifying $Vs_{30}$ values using these techniques. Most research in this area identifies site classes at strong motion stations. Sabegh and Tsang (2011) used artificial neural networks (ANN) to categorize the site class at locations with robust ground motion stations. The investigation utilized both Probabilistic Neural Networks (PNN) and Generalized Regression Neural Networks (GRNN). The study considered four distinct site classes, namely hard rock, rock, stiff soil, and soft soil. As a benchmark for the artificial neural networks employed in the research, four Horizontal-to-Vertical Spectral Ratio (HVSR) curves were used, which were sourced from the works of Ghasemi et al. (2009) and Zhao et al. (2006). The objective was to improve the prognostic abilities of the model while decreasing the burden of data processing. The developed procedure's validation was executed by forecasting the category of 87 robust seismic station sites that had recorded the Mw 7.6 Chi-Chi earthquake in Taiwan in 1999. The PNN method demonstrated a 78% success rate in its predictions, while the GRNN method yielded a 75% success rate in its predictions. The scope of this study was limited solely to site classification rather than identifying specific $Vs_{30}$ values at the station sites. In their subsequent study, Sabegh and Tsang (2014) created a technique for identifying site classes at locations where there is an absence of geological and geotechnical information. A total of 116 recordings were utilized in their study, derived from two earthquakes that occurred on August 11, 2012, in the northwestern region of Iran. The magnitude of these seismic events measured Mw 6.4 and Mw 6.3. Furthermore, the study incorporated conventional pattern recognition techniques from prior literature alongside the previously utilized probabilistic artificial neural networks. Site classes were calculated by applying weights to the results obtained from the different techniques. A site class map was created by extrapolating results between the stations. Sabegh and Rupakhety (2020) utilized the procedure previously developed in Sabegh and Tsang (2011) to classify 60 strong motion station sites in Iran. These stations recorded the Mw 7.3 earthquake that occurred in 2017. HVSR curves for station sites were calculated using these records, and, similar to the previous study by Sabegh and Tsang (2011), HVSR curves from the studies of Ghasemi et al. (2009) and Zhao et al. (2006) were used as references for artificial neural networks. The site classes predicted using AI methods were relatively successful. The studies summarized until now were restricted to predicting site classes rather than estimating $Vs_{30}$ values.

Güllü (2013) utilized ANN and Genetic Expression Programming (GEP) methods by seismic data collected from earthquake monitoring stations located in California using the Pacific Earthquake Engineering Research Center (PEER)–NGA database. The $Vs_{30}$ distribution used in the study isn't explicitly mentioned, although the test results presented show that most sites have a measured $Vs_{30}$ between 400m/s - 600m/s. This database encompasses recorded data of shallow crustal earthquakes in active tectonic regions worldwide. The study compiled earthquake magnitudes, source-to-receiver distances, descriptions of the preferred $Vs_{30}$ at each site, peak ground acceleration values, and pseudo-spectral accelerations (5% damping), among other attributes. A total of 60 recording stations were utilized exclusively within the California region. Between 1952 and 2003, these stations collectively captured a dataset comprising 84 strong ground motion records. Based on the results of the GEP method, the study developed an equation for estimating $Vs_{30}$ at strong ground motion stations within the study area.

The studies mentioned above share a fundamental limitation. The techniques employed in these studies involved feeding accelerometer data into artificial neural networks utilizing probabilistic or generalized regression with limited input data. The methods do not incorporate architectures capable of extracting deep features, which are now considered a common characteristic of modern deep learning approaches. One of the key characteristics that set deep learning apart is its capacity to derive profound and conceptual attributes from signals accomplished through the utilization of extensive databases. In the next sections, we present our study that uses a large data set and employs deep learning approaches. We first present our data set followed by our methodology.

## 3. Data Set

The earthquake recordings used in this study belong to strong motion stations located in Turkey and are operated by the Turkish Disaster and Emergency Management (AFAD). In Turkey, out of the 799 stations operated by AFAD, 594 have available $Vs_{30}$ information measured from field experiments. The data base used in this study comprises 36,417 recordings, (one recording represents signals in 3 different directions) between the years 2012 and 2018. Among these 36,417 recordings, $Vs_{30}$ values are available for 13,974 recordings. The magnitude values of earthquake records vary between M2.2 and M6.5. Figure 1 displays the locations of the strong motion stations along with their corresponding $Vs_{30}$ values and site classes categorized according to NEHRP site classes. Among these site classes, Site Class C represents the highest number of stations, accounting for 57% of all stations. Site Class D corresponds to 30.5%, while Site Class B represents 11.7%. Recordings at stations that fall into site class C accounts for 54.5% of the total recordings used for $Vs_{30}$ prediction, while 29.36% belongs to site class D stations, and 15.79% corresponds to site class B stations.

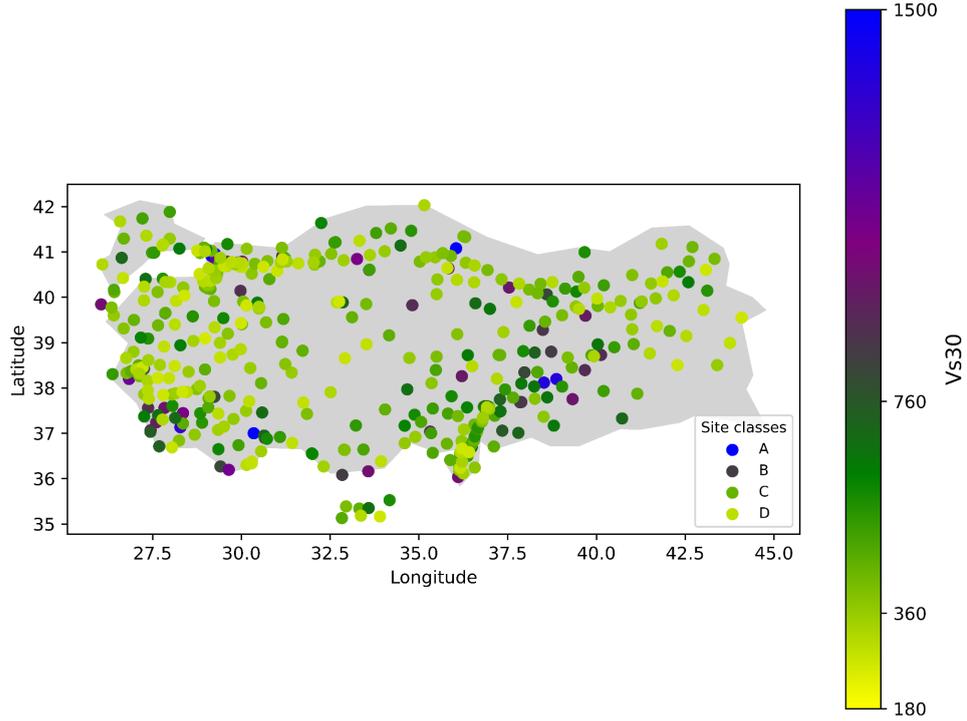

*Figure 1. Local site classes at the AFAD strong motion stations*

## 4. Methodology

In this section, we provide a concise overview of the models that were used, as well as a detailed explanation of the training and parameter optimization phases. We developed two neural network architectures that can effectively analyze accelerometer waveforms in the time and frequency domains.

**4.1 Models**

The structure of both models is shown in Figure 2. Each model consists of an encoder, a decision layer, and an output layer that provides the $Vs_{30}$ value of the event. The encoder uses either ResNet or TCN to capture seismic features. For the decision layer, we chose two fully connected dense layers, to which we add the relative location of the receiver to the source.

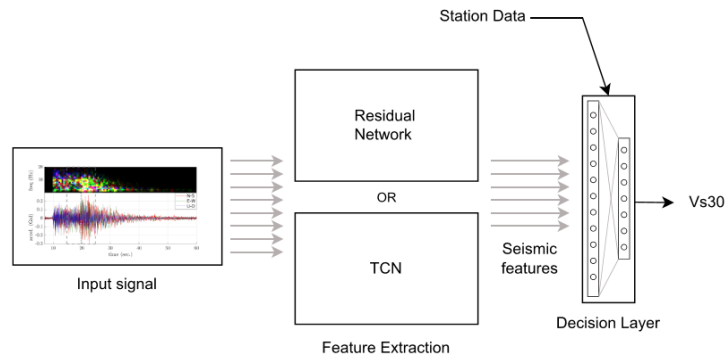

*Figure 2. The architectures of the implemented residual and TCN-based networks*

The following sections provide a detailed explanation of the networks used and the optimization procedures conducted in a more detailed manner.

**4.1.1 Residual Network**

Initially, we employ an encoder that is grounded on the ResNet framework developed by He and Zhang (2015). This framework has been previously utilized in the analysis of seismic waveforms, as shown by Ristea and Radoi (2022). In our current study, we enhance this architecture by introducing a new layer model that is specifically designed to extract distinctive features from individual station recordings. The encoder is composed of convolutional and residual blocks, which operate on an input volume of dimensions Dx51x3 in the frequency domain or DxSx1 in the time domain where D and S indicate the signal dimensions.

To enhance our understanding of the correlation between $Vs_{30}$ and the characteristics extracted from signals in both the time and frequency domains, we incorporated specific max-pooling layers to modify signal dimensions in these domains. These crucial modifications effectively enabled us to train our models, accessing their predictive abilities and ultimately enhancing our capabilities in the fields of earthquake engineering and seismic hazard assessment.

In order to effectively account for the diverse range of features that contribute to an earthquake, adjustments beyond mere network parameter localization and topology complexity were necessary. In light of this need, we incorporated geographical data into our model, feeding the station coordinates (latitude and longitude) to the initial decision layer. Our objective was to evaluate the model's capacity to dynamically predict input signals by linking them with the various subsurface characteristics of the earthquake source regions. Ultimately, the decision layers would reshape this input into a one-value output regression layer, allowing for the prediction of $Vs_{30}$ values.

### 4.1.2 Temporal Convolutional Networks (TCN)

The TCN architecture incorporates causal and dilational convolutional layers tailored to decipher certain deep features from time-domain signals without being influenced by past temporal dependencies S. M. Mousavi and G. C. Beroza (2020). So, when making a prediction at time t, only the information up to that point is used. In other words, it combines all signals and focuses on extracting features only from the last part of the resulting new signal.

Within the context of our research, we extended the applicability of the TCN model to frequency-domain signals as well. To adapt the TCN structure for the adequate training of frequency signals, which always to be higher compared to their time-domain counterparts, we appended max-pooling layers to the end of the first three skip connections in the TCN architecture. This modification effectively reduced the signal dimensions, rendering them more suitable for training purposes.

### 4.2 Training Strategy

Due to the limited availability of data with $Vs_{30}$ information in our $Vs_{30}$ prediction experiments, we employed two distinct training strategies: single-phase training and two-phase training. During single-phase training, we conducted experiments in which both the encoder and decoder were trained using the data specified in Section 3.. We tested these experiments with varying parameters as described below.

In order to enhance our data collection efforts in the face of limited earthquake signal recordings from stations with $Vs_{30}$ data, we implemented a two-phase training process. This approach was chosen to ensure the accuracy and reliability of our data collection efforts. We conducted experiments by transferring appropriate hyper-parameters and results to test the extraction of $Vs_{30}$ characteristics during a study involving earthquake epicenter predictions based on seismic data collected throughout Turkey. By transferring the weights of the convolutional layers, we anticipate the model to fine-tune itself to the shear-wave velocity prediction problem, using the high-level seismic patterns it learned in the initial training phase. We attempted to determine whether the features derived from our prior experiments, which were focused on predicting the epicenter, coincided with the features obtained in our efforts to predict $Vs_{30}$.

Before conducting the training, we preprocessed the signals into a specific format. In order to observe the contributions of weak motion and attenuation behavior to modeling, we bracketed the signal by 15, 30, and 60-second segments of spectra centered around the PGA, which has a critical impact in revealing the path characteristics and propagation process by being the largest component of motion. Initially, we cropped the 3-channel seismic signal, ensuring that the peak ground acceleration (PGA) point of the new signal coincided with the midpoint of the original signal. This way, we aimed to train the feature-rich portions of the seismic signal, which could provide valuable information. If the length of the signal was less than the signal length required for the experiment, we did not use it.

The cross-validation approach we employed in preparing our experiments differs slightly from the conventional practice. In the general context of previous studies on this topic, the specific details of the conducted cross-validation have not been explicitly outlined, and we consider this to be one of the critical aspects. In our study, we performed cross-validation as follows. When partitioning earthquake signal recordings into training and test sets, we ensure that earthquake recordings from the same station are placed exclusively either in the training set or the test set. If a recording is included in the training set from a station, other recordings measured by the same station will not appear in the test set, and vice versa. However, "recordings" (but not stations) may be included in both the training and test sets. Our rationale behind this was the belief that by not intermixing earthquake records from the same recording station locations in the test and training sets, we could potentially facilitate more effective regional testing. In this manner, we aimed to evaluate the features we were teaching in one region within a different region.

In order to operationalize the aforementioned cross-validation technique, a five-fold cross-validation was utilized, where for each fold a non-intersecting 20% of the records are included in a test set. While doing this the stations that collected the recordings of a test set, are not included in the training set of that fold.

**4.3 Parameter Optimization**

We conducted a thorough investigation into the impact of utilizing time and frequency domain representations in our study, implementing both ResNet and TCN architectures. To train our models, we employed a dataset comprising signals with durations of 15, 30, or 60 seconds for parameterization. During the training process, we applied batch normalization to batches of 64 inputs, and we also introduced dropout with rates of 2% and 5% to the final decision layers to enhance the model's ability to generalize.

We adopted the Adam optimizer as our exploration strategy, with an initial learning rate of $10^{-5}$ and implemented a learning rate decay of 0.9 at epochs 5, 10, 20, extending the training for a total duration of 100, 200, 300 epochs. The training utilized either Mean Squared Error (MSE) or Mean Absolute Error (MAE) as the basis for learning.

# 5. Results

**5.1 Single-Phase Training**

At the outset, our trials were conducted for a duration of 60 seconds. The duration of the signals was established as 60 seconds for both the time domain and frequency domain while incorporating dropout rates of 0.1, 0.3, and 0.5, in addition to learning rates of $10^{-5}$, $10^{-6}$, and $10^{-7}$ that were applied to both ResNet and TCN architectures. The most favorable outcomes achieved with these hyper-parameters for signals lasting 60 seconds are depicted in the subsequent diagrams from Figure 3 to Figure 14. The maps represent the percent error in $Vs_{30}$ estimations of 15, 30 and 60-seconds duration experiments. Upon examination of the error map, it is evident that the predictions for $Vs_{30}$ that fall into site class C category are the most accurate with an absolute mean percentage error of 21.1%.

The reader should note that we are not conducting a site classification based on these observations. Our objective is to predict the $Vs_{30}$ value for an arbitrary site with strong motion recordings. In presenting the results for different categories, we intend to assess how well the model works for different categories of sites. Results show that the success for sites that fall into Site Class C category is considerably higher compared to other site classes. The reason for this is that, as mentioned in Section 3, our dataset has an uneven distribution with the largest number of records belonging to Class C. This fact inevitably introduces a bias in our model towards learning site class C and directly affects our results. We anticipate that by collecting higher sample sets for other site classes, we might observe better results for those categories as well.

In addition, varying signal length, learning rate, dropout rate parameters were examined to enhance the results. Two different types of experiments were conducted in terms of input type in frequency and time domains. Separate tests were conducted using the same hyper-parameters for time and frequency signals, and it was observed that frequency signals provided advantages in some instances while similar results were achieved in others.

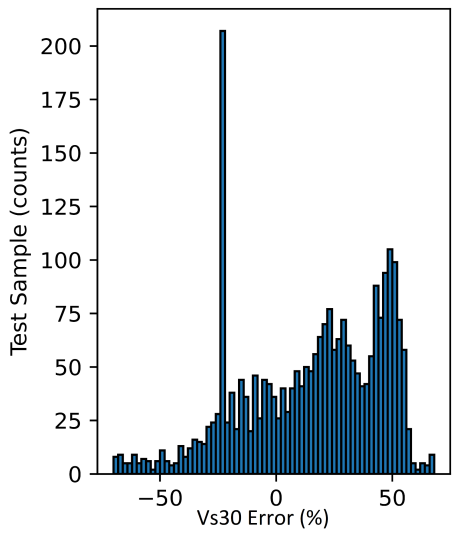
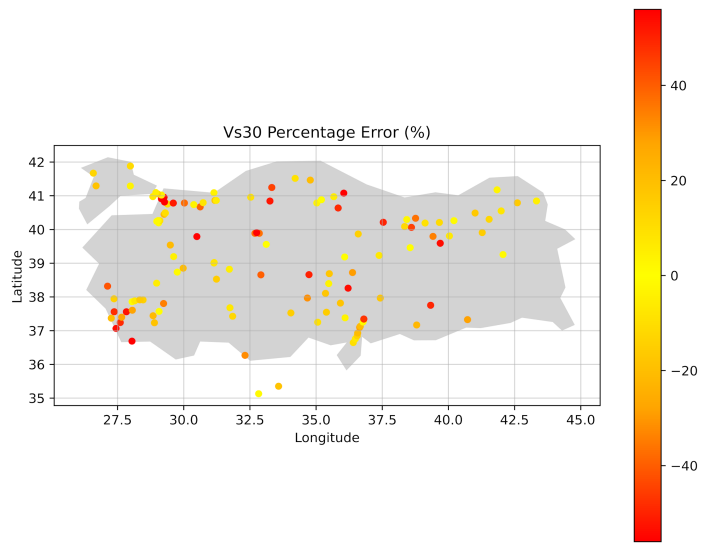

*Figure 3. Vs$_{30}$ Error Histogram for 60-seconds*

*Figure 4. Station-based error map for 60-seconds (%)*

These subsequent figures depict the most favorable outcomes achieved with the hyper-parameters for signals that last 30 seconds. After analyzing the error map, it remains clear that the predictions utilizing site C values exhibit the most accurate results. The absolute average percentage error stands at 26.2%, and for a gain of 30 seconds, there is likely to be an insignificant change in the mean percentage error that can be dismissed.

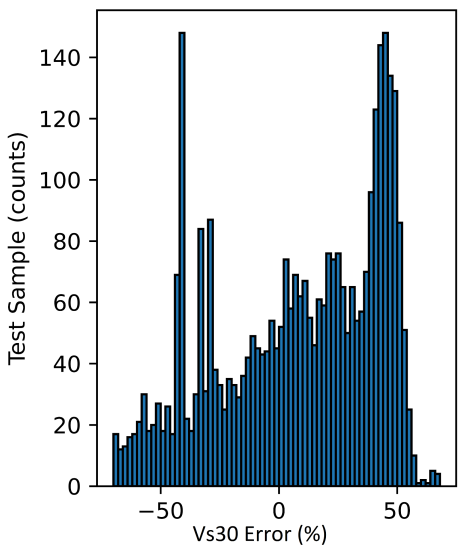
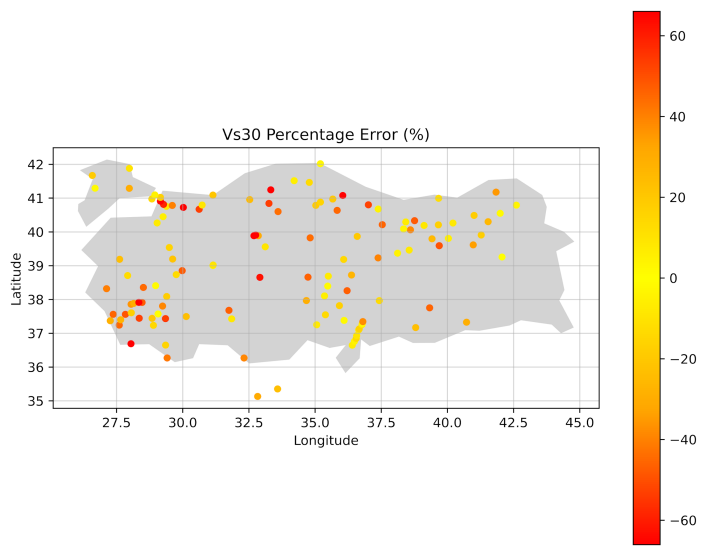

*Figure 5. Vs$_{30}$ Error Histogram for 30-seconds*

*Figure 6. Station-based error map for 30-seconds (%)*

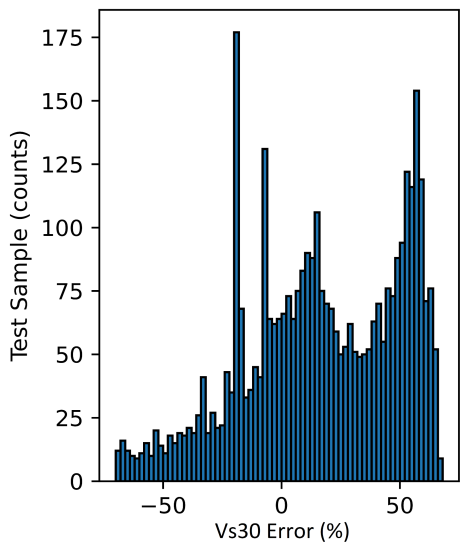
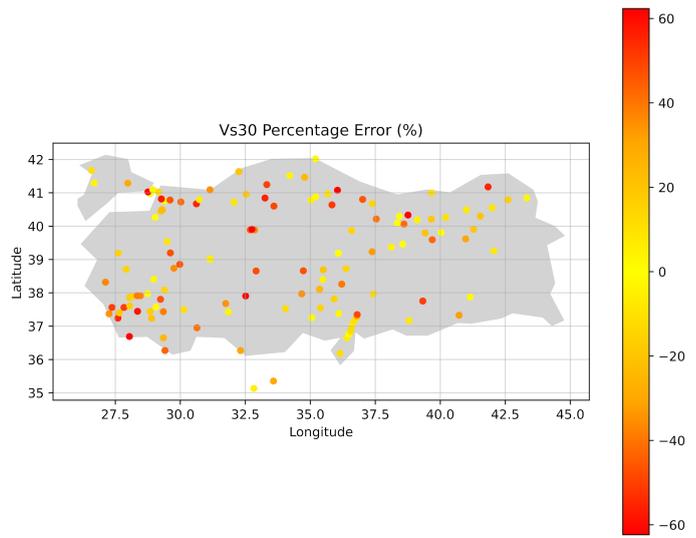

*Figure 7. $Vs_{30}$ Error Histogram for 15-seconds*

*Figure 8. Station-based error map for 15-seconds (%)*

### 5.2 Two-Phase Training

Our objective in this section is to examine the potential advantages of utilizing the high-level characteristics acquired from a previous investigation, which focused on estimating the epicenter of earthquakes using single station recordings. We carried out trials employing signals lasting for 15, 30, and 60 seconds, with the intention of examining whether transferring solely the encoder component of the models that were trained to utilize these signals would have any bearing on $Vs_{30}$ prediction. We transferred this model to estimate $Vs_{30}$ values for the strong motion stations. This two-phase training procedure however, did not produce better results than the single-phase training procedure.

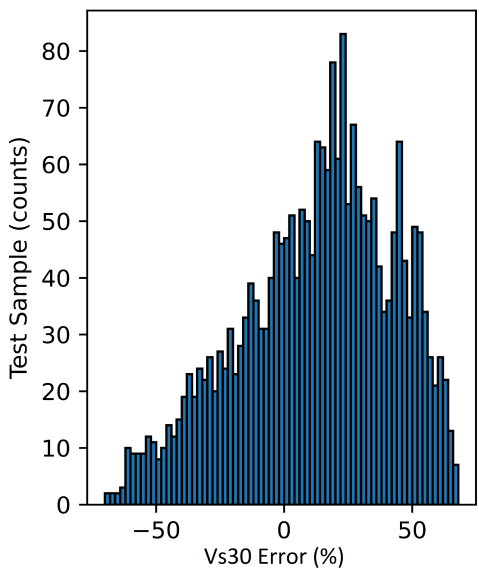
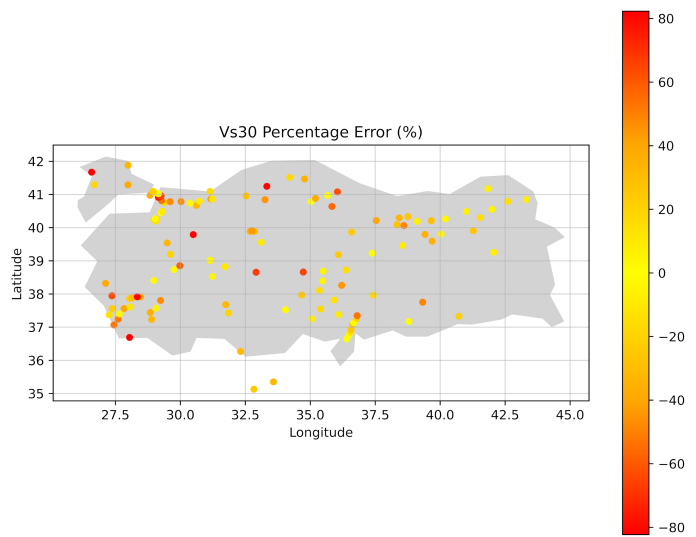

*Figure 9. $Vs_{30}$ Error Histogram for 60-seconds*

*Figure 10. Station-based error map for 60-seconds (%)*

The results of the transfer learning when 30-second duration inputs were used are shown in Figure 11 and Figure 12.

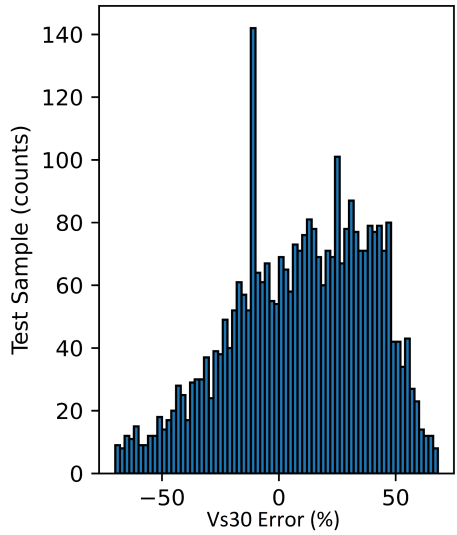
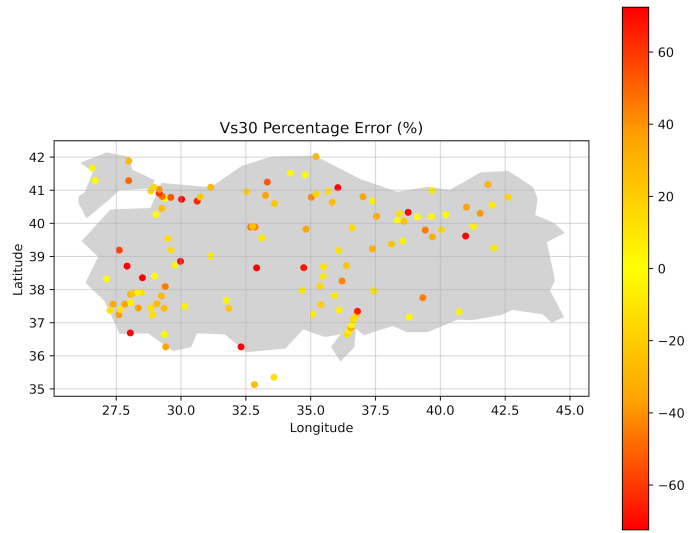

*Figure 11. Vs$_{30}$ Error Histogram for 30-seconds*  *Figure 12. Station-based error map for 30-seconds (%)*

Figure 13 and Figure 14 show the results of transferred experiments conducted with 15-second signals.

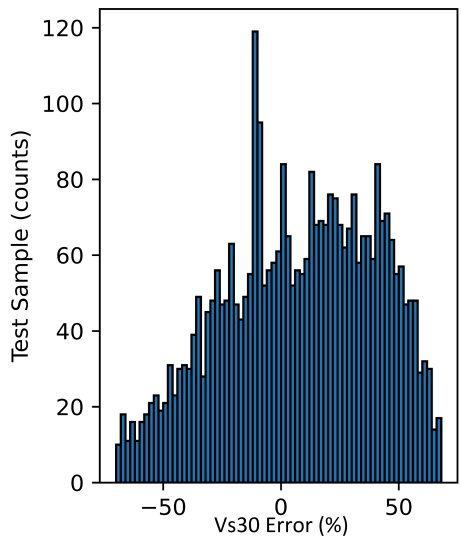
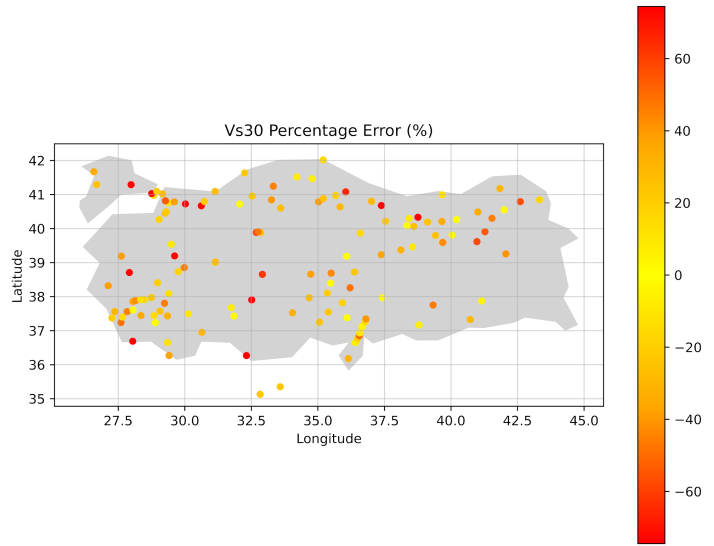

*Figure 13. Vs$_{30}$ Error Histogram for 15-seconds*  *Figure 14. Station-based error map for 15-seconds (%)*

*Table 1 - The Results of Single-Phase Experiments*

|  | 15 Seconds | | 30 Seconds | | 60 Seconds | |
|---|---|---|---|---|---|---|
| Site Class | No. of Stations | Absolute Mean Error | No. of Stations | Absolute Mean Error | No. of Stations | Absolute Mean Error |
| Overall Error | 137 | 23.9% | 129 | 26.2% | 130 | 21.1% |
| A | 1 | 65.4 % | 2 | 70.5 % | 3 | 70.8 % |
| B | 18 | 47.7 % | 19 | 46.9 % | 16 | 49.2 % |
| C | 93 | 13.6 % | 84 | 14.3 % | 90 | 12.3 % |
| D | 25 | 43.5 % | 24 | 47.5 % | 21 | 30.1 % |
| Std 2 | 62.4 % | | 66% | | 56% | |

*Table 2 - The Results of Two-Phase Training Experiments*

|  | 15 Seconds Transfer | | 30 Seconds Transfer | | 60 Seconds Transfer | |
|---|---|---|---|---|---|---|
| Site Class | No. of Stations | Absolute Mean Error | No. of Stations | Absolute Mean Error | No. of Stations | Absolute Mean Error |
| Overall Error | 137 | 30.4 % | 129 | 27.1 % | 130 | 28% |
| A | 1 | 65.4 % | 2 | 66.2 % | 3 | 68.4 % |
| B | 18 | 37.3 % | 19 | 35.9 % | 16 | 45.2 % |
| C | 93 | 22.7 % | 84 | 18.5 % | 90 | 16.3 % |
| D | 25 | 52.5 % | 24 | 46.8 % | 21 | 59.5 % |
| Std 2 | 74.5 % | | 72.4 % | | 82.3 % | |

Despite having relatively good success on predicting $Vs_{30}$ values for some of the stations (<20% mean error on average for some sites), the overall results show that further investigation is required to extract Vs30-related features solely from seismic signals. While our results may not consistently meet the desired level of performance, they provide crucial insights for our future research endeavors. In contrast to prior studies in the literature, we possess a significantly larger dataset with diverse topographical information. This, in turn, offers a more comprehensive perspective for making generalized observations in this domain, although the distribution of our dataset may still not be close to ideal.

As far as comparing results from this study to previous studies goes, the recordings in the test data shared in Güllü (2013), exhibit $Vs_{30}$ values ranging from 400 m/s to 600 m/s. This indicates that at least 20% of the existing strong motion data is confined to a fairly limited range, suggesting a narrow distribution of the dataset. This led us to believe that generalization through such experiments and hence reliable results cannot be obtained through this strategy. To achieve a more comprehensive and reliable outcome, we conducted our experiments using a greater number of stations with $Vs_{30}$ values, including those at the extreme ends of the spectrum. We also ensured a well-balanced distribution of this data throughout both the training and testing phases in our study. In the study conducted by Sabegh and Tsang (2014), 60 stations were used, just like in the Güllü (2013) study, were also utilized, and measurements were based solely on a single earthquake event. For the same reasons, we deemed it unlikely to achieve a robust generalization.

Furthermore, our study included a customized design for cross-validation. As mentioned in the training strategy section, we partitioned the train and test sets in such a way that they did not contain any records from the same station. Our intention in doing so was to perform training in one region and testing in another, thereby avoiding the reuse of the regions we trained on during testing, thus properly cross validating the entire set. However, we did not find any explanation in previous studies regarding the specific cross-validation methodology employed.

We also conducted experiments to investigate the possibility of transferring important features learned from the centroid of a seismic event. When we attempted to transfer the deep features obtained from our experiment with a larger dataset, where we aimed to predict the epicenter location, to the experiment where we were attempting to predict the $Vs_{30}$ value, we did not observe any improvement in the results.

## 6. Conclusions and Future Directions

As seismic waves propagate from the fault to the Earth's surface, they interact with different layers of the Earth affecting their speed, amplitude and frequency content. One important contributing factor to these changes is the dynamic properties of the local site soils. Soils may amplify or deamplify earthquake waves as well as alter their frequency content. Our efforts are currently focused on gaining a deeper understanding of this phenomenon through deep learning methods. We conducted numerous experiments with ResNet and TCN architectures using signals in the frequency and time domain. In general, we obtained better results when utilizing the ResNet architecture and frequency domain signals.

Our investigation indicates that an increase in the number of stations used positively affects the outcomes. Nevertheless, we have observed that the results tend to stabilize and converge towards the stations that fall into site class "C". Two crucial factors arise from this examination. At first glance, it seems that having a greater number of stations ultimately results in more favorable outcomes. This finding may suggest that insufficient data could potentially hinder progress. Furthermore, it should be noted that the quantity of recordings from stations with a site class C has the highest percentage in the overall dataset. Hence, this likely has an effect on the overall error being less for these sites. As more samples are collected for all class sites, we believe that this problem will be alleviated.

In conclusion, by focusing on instances where seismic signals are the strongest, we have demonstrated the potential of deep learning-based approaches in the fields of seismology and earthquake engineering to address complex problems. The findings have instilled hope in us for forthcoming research endeavors. Our observations indicate that studies with a greater abundance of $Vs_{30}$ data tend to yield superior results. Additionally, it may be beneficial to contemplate the implementation of the LSTM architecture in relation to the time signal from an architectural standpoint. LSTMs have the capability to detect long-term relationships and are useful in identifying important characteristics in seismic signals, particularly in data that is sequential. Rather than relying on Fully Connected layers, attention mechanisms can be utilized instead. These mechanisms are becoming more prevalent in modern architecture and have demonstrated favorable outcomes in the majority of cases.

## 7. Acknowledgement

This work is funded by The Scientific and Technological Research Council of Turkey (TÜBİTAK) as a part of our ongoing TÜBİTAK 1001 project, Project No.121M732, titled "Deep Learning and Machine Learning Based Dynamic Soil and Earthquake Parameter Estimation Using Strong Ground Motion Station Records".

## References

[1] National Research Council. (2006). *Improved Seismic Monitoring Improved Decision Making Assessing the Value of Reduced Uncertainty.* The National Academies Press: Washington, D.C.
[2] Molnar, S., Braganza, S., Farrugia, J., Atkinson, G., Boroschek, R., & Ventura, C. (2017). *Earthquake Site Class Characterization of Seismograph and Strong-Motion Stations in Canada and Chile.* In Proceedings of the 16th World

Conference on Earthquake Engineering, Santiago Chile, January 9th to 13th, 2017.

[3] Yaghmaei-Sabegh, S., & Tsang, H. H. (2011). *A New Site Classification Approach Based on Neural Networks*. Soil Dynamics and Earthquake Engineering, 31(7), 974–981.

[4] Wald D. J. ve . Allen T.I., 2007. *Topographic Slope as a Proxy for Seismic Site Conditions and Amplification*, Bulletin of the Seismological Society of America, Vol. 97, No. 5, pp. 1379–1395,

[5] Hassani, B., Atkinson G.M., (2016). *Applicability of the Site Fundamental Frequency as a Vs30 proxy for Central and Eastern North America*, Bulletin of the Seismological Society of America, Vol. 106, No. 2, pp.653–664.

[6] Ni, S., Li, Z., & Somerville, P. (2014). *Estimating subsurface shear velocity with radial to vertical ratio of local P waves*. Seismological Research Letters, 85, 82–90.

[7] Kim, B., Hashash, Y. M. A., Rathje, E. M., Stewart, J. P., Ni, S., Somerville, P. G., Kottke, A. R., Silva, W. J., & Campbell, K. W. (2016). *Subsurface Shear Wave Velocity Characterization Using P-Wave Seismograms in Central and Eastern North America*, Earthquake Spectra, 32(1), 143–169.

[8] Alimoradi, A., & Beck, J. L. (2015). *Machine-Learning Methods for Earthquake Ground Motion Analysis and Simulation.* Journal of Engineering Mechanics, 141(4).

[9] Mosher ve Audet (2020) Mosher, S. G., Audet, P. (2020). *Automatic detection and location of seismic events from time-delay projection mapping and neural network classification.* Journal of Geophysical Research: Solid Earth, 125, e2020JB019426.

[10] Ghasemi, H., Zare, M., Fukushima, Y., & Sinaeian, F. (2009). *Applying Empirical Methods in Site Classification, Using Response Spectral Ratio*. (H/V): A Case Study on Iranian Strong Motion Network (ISMN)." Soil Dynamics and Earthquake Engineering, 29(1), 121–32.

[11] Zhao, J. X., Irikura, K., Zhang, J., Fukushima, Y., Somerville, P. G., Asano, A., et al. (2006). *An Empirical Site-Classification Method for Strong-Motion Stations in Japan Using H/V Response Spectral Ratio.* Bulletin of the Seismological Society of America, 96(3), 914–25.

[12] Sabegh, Yaghmaei-Sabegh, S., & Tsang, H. H. (2014). *Site Class Mapping Based on Earthquake Ground Motion Data Recorded by Regional Seismographic Network.* Natural Hazards, 73, 2067–2087.

[13] Sabegh Rupakhety, 2020 Yaghmaei-Sabegh, S and Rupakhety R, *A new method of seismic site classification using HVSR curves: A case study of the 12 November 2017 Mw 7.3 Ezgeleh earthquake in Iran*, Engineering Geology, 270(5), 2020,

[14] Güllü, H. (2013). *On the Prediction of Shear Wave Velocity at Local Site of Strong Ground Motion Stations: An Application Using Artificial Intelligence.* Bulletin of Earthquake Engineering, 11, 969–997.

[15] K. He, X. Zhang, S. Ren, and J. Sun, *Deep residual learning for image recognition,* 2015

[16] N.-C. Ristea and A. Radoi, *Complex neural networks for estimating epicentral distance, depth, and magnitude of seismic waves*, IEEE Geoscience and Remote Sensing Letters, vol. 19, pp. 1–5, 2022.

[17] S. M. Mousavi and G. C. Beroza, *Bayesian-Deep-Learning Estimation of Earthquake Location From Single-Station Observations,* in IEEE Transactions on Geoscience and Remote Sensing, vol. 58, no. 11, pp. 8211-8224, Nov. 2020, doi: 10.1109/TGRS.2020.2988770.